\pdfoutput=1
\documentclass{article}
\usepackage{arxiv}
\usepackage{amsmath,amssymb,amsfonts}
\usepackage{algorithmic}
\usepackage{graphicx}
\usepackage{textcomp}
\usepackage{cite}
\usepackage{url}

\title{From Creation to Curriculum: Examining the role of generative AI in Arts Universities}

\author{
Atticus Sims\\
Faculty of Media Creation\\
Kyoto Seika University\\
Kyoto, Japan\\
atticus@kyoto-seika.ac.jp
}

\begin{document}

\maketitle
\begin{abstract}
  The age of Artificial Intelligence (AI) is marked by its transformative "generative" capabilities, distinguishing it from prior iterations. This burgeoning characteristic of AI has enabled it to produce new and original content, inherently showcasing its creative prowess. This shift challenges and requires a recalibration in the realm of arts education, urging a departure from established pedagogies centered on human-driven image creation. The paper meticulously addresses the integration of AI tools, with a spotlight on Stable Diffusion (SD), into university arts curricula. Drawing from practical insights gathered from workshops conducted in July 2023, which culminated in an exhibition of AI-driven artworks, the paper aims to provide a roadmap for seamlessly infusing these tools into academic settings. Given their recent emergence, the paper delves into a comprehensive overview of such tools, emphasizing the intricate dance between artists, developers, and researchers in the open-source AI art world. This discourse extends to the challenges and imperatives faced by educational institutions. It presents a compelling case for the swift adoption of these avant-garde tools, underscoring the paramount importance of equipping students with the competencies required to thrive in an AI-augmented artistic landscape.
  \end{abstract}
  
  \section{Introduction}
  
  It can be contended that we currently find ourselves in the age of Artificial Intelligence (AI). While AI has seamlessly integrated into various facets of our daily lives over the years, the distinct characteristic that sets the present wave apart is its "generative" nature. Instead of merely cataloging and organizing information as previous iterations might have, contemporary AI has the capability to synthesize and produce novel information, making it inherently 'creative'. This burgeoning shift not only introduces an innovative technical methodology for image production but also necessitates a paradigmatic rethinking in the domain of arts education. As AI delves into realms previously reserved for human imagination, it challenges traditional pedagogies and conceptual frameworks surrounding the process of image creation.
  
  This paper seeks to address the integration of AI tools in university arts education. In order to explore this topic at a practical level, a series of workshops were held in July 2023 which led to a group exhibition of AI generated works created by participants in the workshops. These activities focused on understanding the techniques and processes of creating images with Stable Diffusion, an open-source generative AI tool that provides much greater control in determining the final form of an image. These workshops and subsequent one-on-one guidance of students using these technologies will be examined in the third section in order to explore possible ways in which these tools can be incorporated into university classrooms and curriculua. 
  
  As these tools are a recent development and represent a drastically different approach to image creation, an overview of the technology behind these tools as well as the most important technical aspects of their operation will be described in detail. First, an overview of the key terms and concepts behind AI image generation will be presented with the aim of providing non-technical readers with an intuitive understanding of how this software works. Additionally, a brief analysis of the drivers behind the continued advancement and increasing sophistication of such tools will be presented, focusing on the dynamic interplay between artists, developers and researchers within the open-source AI art ecosystem. Next, we will examine the specifics of generating images with Stable Diffusion, including core concepts such prompting, models, seed values, samplers and VAE, as well as extensions to the base Stable DIffusion workflow that have become valuable parts of the AI artist repertoire such as image-to-image, inpainting, ControNet, upscaling and supplementary custom models (usually referred to a LoRA).
  
  As this technology has faced a great deal of criticism and controversy, I will address some of these issues including the debate over originality of AI artworks and copyright concerns about the training of the models. In addition to these concerns, there has also been much discussion regarding the impact these tools will have within creative industries, and I will briefly address these issues by presenting how these tools are already impacting various industries and then assessing the potential of these tools to transform media in both positive and negative ways.
  
  Finally, I will provide some potential top-down strategies for arts universities to incorporate these rapidly evolving tools into their curricula, as well as some steps that individual teachers can take to better understand and utilize these tools in their classrooms.

  \section{Overview of generative AI art tools}
  
  In this section, we will provide a basic overview of the generative AI art tools for image creation, briefly examining the of the nature of generative AI, followed by an overview of the technology behind AI image generation tools, an analysis of the open-source ecosystem that is driving generative AI art, and a practical survey of the key components of Stable Diffuion.
  
  \subsection{What is generative AI?}
  Artificial Intelligence (AI) is the capability of a machine to imitate intelligent human behavior. It encompasses systems or machines that can perform tasks that usually require human intelligence, such as visual perception, speech recognition, decision-making, and language translation.
  
  AI, as opposed to traditional computer programming, is based on the concept of learning rather than providing step-by-step instructions to the computer. AI programs are often referred to as ‘models’ which learn to identify patterns in the data in a process called ‘training’. Once a model is trained, it can then make predictions or decisions based on data it has not seen before. This process is called ‘inference’. This is analogous to a radiologist being trained to recognize patterns in medical images from a text book, then applying this pattern recognition ability in clinical settings with medical images they have never seen before.
  
  AI models can be broken down into two main categories: discriminative models and generative models. Discriminative models, as the name suggests, discriminate certain patterns in data. Examples of this are facial recognition software or optical character recognition. Generative models on the other hand, are trained to produce novel data that is similar to the training data. It can be said that the goal of generative models is to “build a model that can generate new sets of features that look as if they have been created using the same rules as the original data.” An example of this is instructing and image generation model to create an image of an astronaut riding a horse in the style of cubism. The result will be an image that contains the key aesthetic features of cubism while being a novel composition.
  
  Generative models are probabilistic rather than deterministic, and they require a stochastic element that influences the outcome. This is often referred to as the “seed” or “random seed”, which will be discussed in greater detail below\cite{foster2022generative}. 
  
  The recent boom in AI is due primarily to advancements in generative modelling, specifically breakthroughs in large language models (LLM) such as OpenAI’s GPT models. LLMs are generative models because they are trained on a large dataset (text on the internet, etc.) and produce new data that is similar to the training data, i.e., natural language. Since these models are trained on an enormous body of text produced by humans, we are able to interact with them using natural language.
  
  Generative models for image creation such as Generative Adversarial Networks (GAN) and Variational Autoencoders (VAE) have been around for a number of years, but the recent advancements in AI image creation were enabled through the integration of generative language models such as GPT. Image generation models such as Stable Diffusion differ from previous models such as GAN due to the ability to take a text description of an image and create an image that matches the verbal description. This process is known as text-to-image (TTI or t2i) and is the foundation of all commonly used AI image software in use today.

  \subsection{Overview of the technology behind AI image generation tools}
  
  \subsubsection{Early attempts at TTI}
  Text-to-image (TTI) models are a relatively new area of research in the field of computer vision and artificial intelligence. The first notable attempt at text-to-image synthesis was in 2014 by a team of researchers from the University of Montreal, who proposed a model called Generative Adversarial Networks (GANs), which could generate images from textual descriptions\cite{goodfellow2020generative}.  Although GANs are not technically considered TTI models, there were notable applications of GANs in early attempts at text to image generation such as the Deep Convolutional GANs (DCGANs)\cite{radford2015unsupervised}, and Attention-based GANs (AttnGANs)\cite{Xu2017AttnGANFT}.
  
  Another notable advancement in image generation is Variational Autoencoder (VAE), which plays a significant role in latent diffusion models (see Fig. 3). The core concept behind VAE is that data, for example, pixel values from an image, are encoded into a mathematical representation of the image, often referred to as "latent space", and then decoded to reconstruct the original image (Fig. 1). For a more detailed technical description see \cite{cinelli2021variational} \cite{Kingma2013AutoEncodingVB} \cite{Higgins2016betaVAELB}.
  
  \begin{figure}[htbp]
  \centerline{\includegraphics[width=1.0\columnwidth]{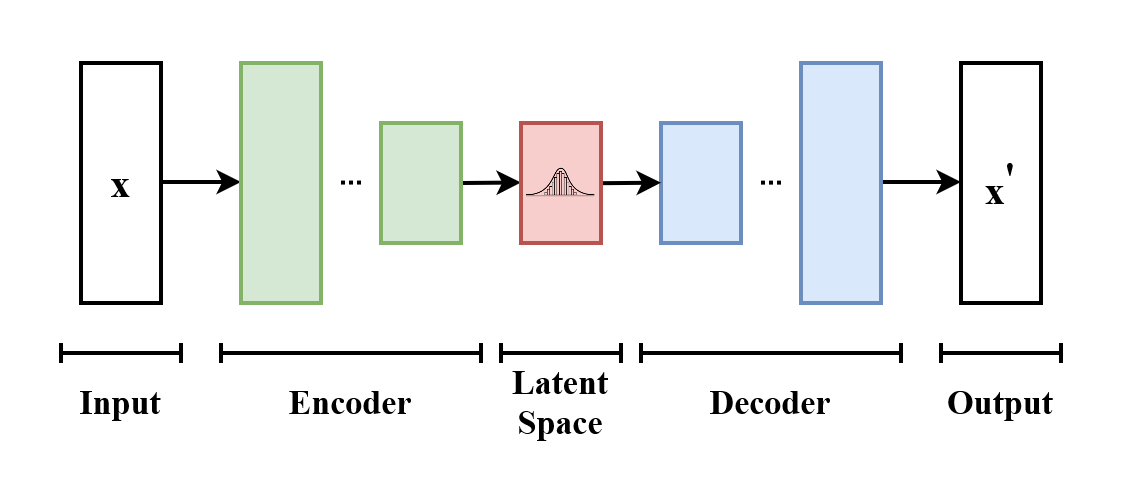}}
      \caption{Simplified visual representation of VAE architecture \cite{VAEBasic89:online}}
      \label{fig:enter-label}
  \end{figure}
  
  \subsubsection{Diffusion Models}
  In 2021, OpenAI introduced a new approach to text-to-image synthesis using the diffusion model which was first introduced in 2015 \cite{SohlDickstein2015DeepUL}. A diffusion model is trained to generate high-quality images by iteratively "diffusing" the image with Gaussian noise through a sequence of discrete time steps. During each time step, the model applies a series of invertible transformations to the image, which results in a sequence of progressively "noisier" representations. When used to generate novel images, a random noise image is used as the "seed" and the model then  generates a sequence of images that become increasingly sharper and more detailed a certain number of "steps", resulting in an image that matches the input specifications, which in the case of TTI is the 'prompt'. \cite{Croitoru2022DiffusionMI} \cite{Yang2022DiffusionMA}.
  This process can be intuitively understood by observing Fig X below. During training, noise is iteratively introduced to an image until a 'noise' image is achieved. This is represented as the "Fixed Forward Diffusion Process" in the figure. When the trained model is used to generate images, the model is provided with a noise image, the 'seed', and noise is reduced in a series of steps. This is represented in the figure as "Generative Reverse Denoising Process". The key idea is that through the denoising process the model will seek to match the noisy image with the prompt. A loose analogy is that the visual phenomenon of pareidolia, which is the tendency to see meaningful images in random patterns such as clouds\cite{Pareidol35:online}. 
  
  The diffusion model has several advantages over previous types of models, including better stability and the ability to generate higher quality images with more detail and fewer artifacts\cite{Dhariwal2021DiffusionMB}. However, it also has some limitations, such as being computationally intensive and requiring a larger amount of training data\cite{Rombach2021HighResolutionIS}.
  
  \begin{figure}[htbp]
  \centerline{\includegraphics[width=1.0\columnwidth]{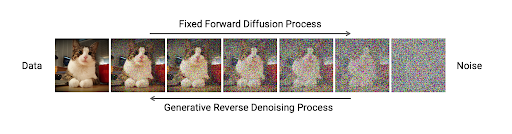}}
      \caption{Diffusion Process \cite{DiffusionImage1:online}}
      \label{fig:enter-label}
  \end{figure}
  
  \subsubsection{CLIP}
  A key component in the TTI revolution is the introduction of Contrastive Language-Image Pre-Training (CLIP), a neural network model developed by OpenAI in 2021 that can understand natural language text and images in a joint embedding space\cite{CLIPConn49:online}. Simply put, a joint embedding space is like a meeting point where different types of data such as pictures and words are turned into similar formats so they can be easily compared or matched. The CLIP model is pre-trained on a large dataset of images and their associated captions, allowing it to learn to associate textual descriptions with visual concepts and features \cite{Ramesh2021ZeroShotTG}.
  
  Its purpose is to enable machines to understand natural language in the context of visual information, allowing them to perform tasks such as image classification, image retrieval, and text-to-image synthesis.\cite{Dosovitskiy2020AnII}\cite{Galatolo2021GeneratingIF}.  In text-to-image synthesis, CLIP plays a critical role in enabling models to generate images that match a given textual description. One of the unique features of CLIP is its ability to perform cross-modal retrieval, which means it can retrieve images that are semantically similar to a given textual description and vice versa. This makes CLIP highly versatile and applicable to a wide range of tasks beyond text-to-image synthesis, including image classification, image retrieval, and visual question answering\cite{Ramesh2022HierarchicalTI}
  
  The use of transformers, the architecture used in CLIP, is a significant contributor to the recent advancements in computer vision with significant advantages over convolutional neural networks utilized in GANs. Distinguishing features of transformers include bidirectional feature encoding and a capacity for large-scale pre-training, allowing them to process multiple modalities such as text, audio, images and video\cite{Khan2021TransformersIV}, In addition to the advances in transformer architecture, the release of CLIP source code\cite{openaiCL34:online} by OpenAI under the MIT License has been fundamental in the current explosion of generative media software. 
  
  \subsubsection{Latent Diffusion Models}
  The CompVis group at LMU Munich introduced the latent diffusion model (LDM) in a paper published in December 2021\cite{Rombach2021HighResolutionISpreprint}. In contrast to previous diffusion models, the LDM works in latent space\cite{Kingma2013AutoEncodingVB}\cite{radford2015unsupervised} instead of pixel space. This means that instead of generating images directly as pixels, the LDM is trained to encode raw data such as images and text into compressed mathematical representations (latent space). The diffusion process is then carried out on these compressed representations, the result of which which is then decoded as pixels. The key advantages of the LDM over previous diffusion models are that it allows for better control over the generated images, it can generate images at higher resolutions with fewer artifacts than previous diffusion models\cite{Rombach2021HighResolutionIS}, and due to the compression of images, the training of these models can be carried out with significantly less computation and results in much smaller model sizes that can be run on consumer hardware\cite{Revoluti9:online}. .
  \begin{figure}
      \centering
      \includegraphics[width=1\linewidth]{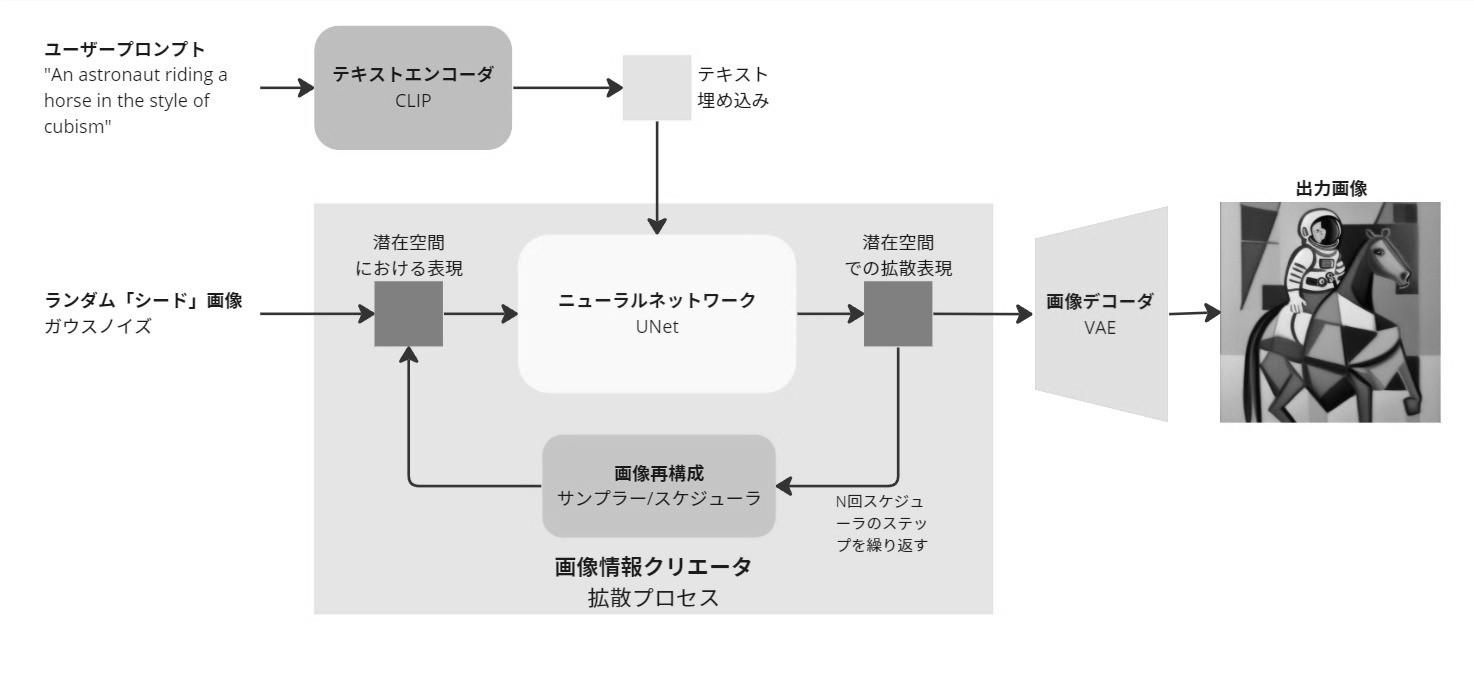}
      \caption{Simplified latent diffusion model architecture}
      \label{fig:enter-label}
  \end{figure}
  
  The original latent diffusion model was further developed and trained on the LAION-5B image dataset\cite{Schuhmann2022LAION5BAO} with the support of Stability.AI, a London-based AI startup\cite{Stabilit37:online}, resulting in significant improvements in image quality and model compression \cite{Stabilit37:online}. The model and source code were publicly released in August 2022 as Stable Diffusion under a Creative ML OpenRAIL-M license, allowing commercial and non-commercial use\cite{StableDi46:online}. This resulted in the spawning of hundreds of model variants and community driven innovations that have been incorporated into open-source implementations of the core model\cite{StableDi68:online} and has seen unprecedented adoption among developers\cite{ArtIsntD24:online}.

  \subsection{The Generative AI Art Ecosystem}
  In the previous section, a technical overview of the development of the primary tools and frameworks utilized in generative AI  image creation was given. These developments represent significant advancements in the fields of computer vision and artificial intelligence that have proven to be transformative in how media is created, and their applications have proven to have a significant impact on society. In the case of artistic and creative media such as image, video and sound, diffusion based models have garnered much attention in the public sphere, focusing both on their capabilities as well as the disruptive effects they are having on artistic production and practices. In this section I will examine the role of community led development of generative art tools, focusing primarily on the open-source development that utilizes Stable Diffusion as its basis.

  \subsubsection{TTI Frameworks}
  The three primary TTI frameworks that have been made publicly available are DALL-E, Midjourney and Stable Diffusion. OpenAI’s DALL-E 2 was opened for research preview in July 2022 and made publicly accessible in September\cite{DALL·E213:online}. This model garnered significant attention due to it being the first TTI capable of producing high quality images from text prompts. OpenAI has not released the source code for DALLE-2, and at present it is only accessible through OpenAI’s website as a paid service. Midjourney is an "independent research lab\cite{Midjourn87:online}" whose TTI service entered open beta in July 2022\footnote{https://twitter.com/midjourney/status/1547108864788553729} and operates on a similar credit subscription business model as DALL-E 2. Training data sets and the underlying TTI model used by Midjourney are unknown, but it is assumed that it shares similar training data and architecture with OpenAI and Stable DIffusion. Image generation in Midjourney is conducted entirely through a discord server though formatted commands. 
  
  Although sites such as these allow for the fast generation of high quality images, these services, like DALL-E 2 and Midjourney, are limited to image generation and lack the breadth of affordances that are available in open-source implementations of Stable Diffusion. For the remainder of this article, I will focus solely on open-source tools that are actively used in the generative AI art communities. 
  
  \subsubsection{Low-code/No-code User Interfaces for still images}
  Prior to the release of Stable Diffusion, there were a number of AI art frameworks that utilized low-code/no-code UIs, primarily Google Colab\cite{Aigenera37:online}. Google Colab is a browser based Python notebook similar to Jupyter, which allows users to write and run Python based machine learning applications that run on Google cloud servers\cite{GoogleColabT63:online}. Due to their notebook format, it is possible to use a shared machine learning Colab notebook with little or no knowledge of coding, and the use of cloud compute enabled artists to use computationally intensive processes (such as GANs) without a powerful computer. Google Colab continues to play a significant role in the creation of AI generated art, especially computationally expensive animation.

  \begin{figure}
      \centering
      \includegraphics[width=0.5\linewidth]{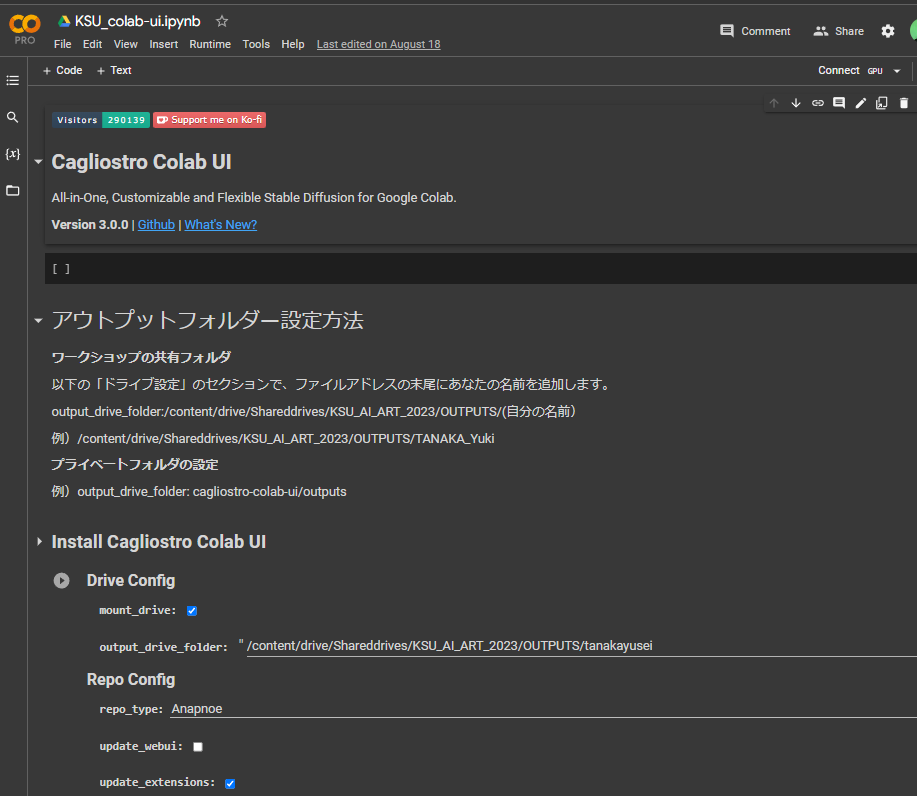}
      \caption{Google Colab Notebook}
      \label{fig:enter-label}
  \end{figure}
  
  In the AI art community, there is a strong connection between artists, developers and researchers. Advancements such as those described previously often have source code that is published to GitHub repositories. Developers utilize these frameworks by adapting them for low-code/no-code use, combining them with other frameworks or integrating them as new features in existing frameworks. As standalone tools these are often showcased as demos on Hugging Face, a platform established in 2016 that focuses on democratizing machine learning\cite{HuggingF26:online}. Hugging Face shares features with both Google Colab and Github, in that it integrates no-code interfaces for running machine learning apps, while simultaneously serving as an open repository for machine learning code bases. 
  
  Another widely used no-code UI in AI art is Gradio, which was developed in 2019 in collaboration with Stanford university to enable non-technical researchers to use machine learning applications\cite{Abid2019GradioHS}. Gradio was acquired by Hugging Face in 2021, and the use of Gradio as a UI for AI art has become standard, especially for Stable Diffusion users running the model locally on consumer-grade hardware (Google Colab only runs on the cloud). One of the most widely used Stable Diffusion implementations is AUTOMATIC1111, which is maintained by a community of over 300 contributors\footnote{It is difficult to accurately estimate the total number of AUTOMATIC1111 users, but the GitHub repository has been starred over 38,000 times.} and is continuously updated with practical implementations of cutting edge research including extensions, samplers, upscalers, etc.(see next section for detailed explanation)\cite{AUTOMATI2:online}.

  \subsection{Key Components of Stable Diffusion}
  
  To grasp the basic operation of Stable Diffusion, it becomes crucial to understand the integral components that steer image generation. For the context of this section, the Automatic1111 user interface serves as our focal point of analysis. However, it's essential to note that the predominant aspects guiding and controlling image generation can be extended to other popular user interfaces, such as ComfyUI. These aspects are intrinsic to the Diffusers Pipeline, the predominant open-source code framework tailored for developing applications based on diffusion-centric generative AI technology.
  
  Delving into the intricacies of Stable Diffusion, it's pivotal to recognize that its foundational architecture is an assemblage of subcomponents, each playing a distinctive role, as depicted in Fig. 3. Each facet of the Stable Diffusion architecture emanated from diverse research endeavors. Furthermore, multiple variants of each component exist, each capable of modifying the final image output in different magnitudes. Although every element of the base architecture offers the potential for manipulation, this article's scope will be confined to those predominantly utilized in artistic workflows.
  
  Beyond these core components, additional controls exist, albeit external to the primary Stable Diffusion architecture. These are usually bundled within user interfaces as default features and can be grouped under two categories: scripts and extensions.
  
  Scripts typically offer rudimentary utilities, enhancing the artist's workflow. A prevalent example is the 'x/y/z plot' script, deployed to assess the impact of distinct parameter configurations.
  
  In contrast, extensions are intricate additions to the principal Stable Diffusion pipeline. They are crafted and sustained independently from both the primary diffusers pipeline and the user interface. Extensions often draw inspiration from avant-garde computer vision research and are finetuned to seamlessly integrate with the Stable Diffusion pipeline and UI. The Automatic1111 GitHub repository chronicles officially endorsed extensions. However, numerous unofficial extensions also permeate the ecosystem. A frequent trajectory observed is the evolution of a potent extension from an optional UI add-on to a default integration, especially if it becomes a staple in the standard artistic process. ControlNet and Extra Networks, both discussed subsequently, epitomize this transition.
  
  \subsubsection{Primary Parameters and Controls}
  
  \paragraph{Checkpoints}
  Frequently dubbed as the 'model', the checkpoint stands out as the pivotal component in AI image generation. Stability AI, the entity behind Stable Diffusion, has unveiled multiple iterations of the Stable Diffusion base model (Table 1). Trained on expansive image datasets like the LAION 5B, these models harness the computational might of extensive GPU arrays. As open-source entities, these base models provide a springboard for developers to craft bespoke models—either by supplementing images or amalgamating existing checkpoints. This has catalyzed the emergence of a plethora of custom models, accessible on platforms like CivitAI. The eclectic range of these models accentuates the allure of open-source software in AI image generation.
  
  \begin{table}
      \centering
  \caption{Image sizes for SD Base Models}
  \label{tab:my_table}
      \begin{tabular}{|c|c|} \hline 
           Base Model Version & Image size 
  \\ \hline 
           SD1.5 & 512x512 
  \\ \hline 
           SD2.1 & 768x768 
  \\ \hline 
           SDXL & 1024x1024 \\ \hline
      \end{tabular}

  \end{table}

  \paragraph{Seed}
  
  At the core of Stable Diffusion lies the concept of the seed value. Each seed value engenders a distinct noise image, foundational to the diffusion process. It's paramount to discern that identical seed values will replicate an image, while variations will yield different outcomes.
  
  \paragraph{Steps}
  
  The 'Steps' setting delineates the iteration frequency of noise image diffusion until the final rendition. Its relevance is intertwined with the Sampling Method discussed subsequently. Generally, an incremental step count elevates image quality, albeit with diminishing returns beyond a specific threshold.
  
  \paragraph{Sampling Method}
  
  Referred to interchangeably as 'sampler' or 'scheduler', the sampling method reconstructs the image post each diffusion iteration. From the artist's perspective, the crux lies in recognizing the variance in image outcomes based on the sampling method chosen.
  
  \paragraph{Prompt}
  
  Prompting remains pivotal to TTI generation. In the context of Stable Diffusion, the emphasis is on iterative experimentation. The Automatic1111 interface also introduces a 'Negative prompt' feature, allowing artists to delineate unwanted elements in the final image. 
  
  Within Stable Diffusion, prompts can be weighted, highlighting their relative significance. This weighting can be achieved using nested parentheses or through a formatted directive like (text:weight), offering granular control over the diffusion process. For example, (happy dog: 0.8) will assign a weight of 0.8 to the phrase ‘happy dog’

  \begin{table}
      \centering
  \caption{Prompt Weight Variations}
  \label{tab:my_label}
      \begin{tabular}{|c|c|c|} \hline 
           Prompt&  Weight (happy)& Weight (dog)\\ \hline 
           happy dog&  1.0& 1.0\\ \hline 
           (happy) (dog:0.5)&  1.1& 0.5\\ \hline 
           (((happy dog)))&  1.3& 1.3\\ \hline 
           (happy dog:1.3)&  1.3& 1.3\\ \hline
      \end{tabular}

  \end{table}

  \subsubsection{Extensions}
  
  Understanding the intricacies of AI image generation necessitates diving into extensions that elevate the capability of base models. Herein, we discuss prominent extensions, their functionalities, and how they empower artists in their creative endeavors.
  
  \paragraph{Extra Networks}
  
  Extra Networks is a collection of three specific fine-tuning methodologies: textual inversion, LoRA, and Hypernetworks. However, due to the predominant use and efficacy of the more recent LoRA models, Hypernetworks will not be covered in this discussion.
  
  Though each of these methodologies has distinct underlying technologies, they all share a common principle. Think of them as supplementary layers appended to the principal checkpoint, influencing the style or subject of a resultant image.
  
  \subparagraph{Textual Inversions}
  
  Often termed 'embeddings', textual inversions are unique trigger words predominantly used as negative prompts. Their primary function is to suppress undesired attributes within the generated image.
  
  \subparagraph{LoRA}
  
  An acronym for Low Rank Adaption, LoRA's origins can be traced back as a technique developed to fine-tune substantial language models\cite{hu2021loralowrankadaptationlarge}. Its adaptation to the realm of image diffusion has been transformative, positioning it as a worthy successor to the less efficient Hypernetworks. Fundamentally, the training of LoRA instills new weights into a base model without tampering with its original structure. A pivotal aspect of LoRA is its compatibility. A LoRA model honed on a specific Stable Diffusion base model seamlessly integrates with any custom model under the same category. However, cross-category compatibility is absent.
  
  Training a LoRA model is quite flexible. While as few as 10 images can suffice, a range of 30 to 300 images is typically recommended. Other merits of LoRA include modest processing power requirements and swift training times. Such attributes have cemented LoRA's position as a quintessential tool for AI artists, as evidenced by the plethora of LoRA models on platforms like CivitAI.

  \begin{table}
      \centering
  \caption{Categories of LoRA}
  \label{tab:my_label}
      \begin{tabular}{|c|c|} \hline 
           \textbf{Type}& \textbf{Example}\\ \hline 
           style& abstract expressionism, claymation, Renoir, line drawing, cyberpunk\\ \hline 
           aesthetic& white, red, fog, underwater\\ \hline 
           subject& orchids, robots, Brad Pitt, neon lights\\ \hline
      \end{tabular}

  \end{table}
  
  LoRA's versatility is evident in its broad categorizations: style, aesthetic, and subject. When initiating a LoRA, specific prompt formats, often inclusive of weight assignments, are employed. Additionally, certain trigger words may be mandated. An intriguing facet is the potential to amalgamate multiple LoRAs within a singular image (Table 3) to mix multiple styles, aesthetics or subjects in unique ways.
  
  \paragraph{ControlNet}
  
  Introduced in February 2023, ControlNet is a groundbreaking neural network architecture tailored for spatial conditioning control within diffusion models. It seamlessly amalgamates various computer vision techniques, from depth and edge detection to pose estimation and object segmentation, offering artists unprecedented control over image composition\cite{zhang2023addingconditionalcontroltexttoimage}.
  
  Before ControlNet's advent, artists relied on diverse tactics for directing image composition. ControlNet's emergence reshaped this landscape, offering artists a streamlined, potent mechanism for image composition guidance.
  
  In essence, both ControlNet and LoRA represent monumental advancements in the AI image generation sphere. They have equipped AI artists with an unparalleled degree of compositional and stylistic control, particularly in a domain characterized by its inherent randomness and unpredictability.

  \section{AI image generation tools and education}
  
  \subsection{A Constructionist Perspective on Learning Generative AI Tools: A Comparative Analysis with Photography}
  
  This section elucidates the learning process of generative AI tools through the lens of Seymour Papert's constructionist theory of learning and education\cite{Construc31:online}. By juxtaposing AI image creation with the well-established medium of photography, I aim to illuminate the unique affordances and constraints that AI offers in the context of artistic production.
  
  Generative AI tools, as highlighted in the technical overview section of this paper, are inherently complex. Nevertheless, with the advent of low-code/no-code platforms like Google Colaboratory and Gradio, an in-depth grasp of the underlying technology isn't a prerequisite. Taking a cue from Papert’s constructionism—succinctly distilled as "learning by making"—these tools offer a rich environment for learners. The rapid and iterative prototyping capabilities enable students to form mental models about the software's functionality and beyond.
  
  More than a mere technical utility, these tools offer a deep dive into the nuances of "image" as a concept. Students can delve into aesthetic dimensions like composition, color, and space, afforded by the iterative creation processes embedded within these platforms. In many respects, these tools bear a closer resemblance to photography than traditional mediums like painting. The feedback loop in photography, spanning from conceptualization ("That could be an intriguing photo") to realization (seeing the captured image), is more instantaneous than the time-intensive process of painting. 
  
  Photography allows for prolific creation, where artists often sift through hundreds of images to select the "final work." This pattern can be even more pronounced in the realm of AI artistry. Take, for instance, the multifaceted settings of a camera—from f-stop to pixel density—which can modify the resultant image in manifold ways. These parameters, although intricate, become intuitively understood by photographers over time, a mastery achieved through repetitive usage and the development of mental models.
  
  Such experiential learning can be witnessed in photography, where photographers, by shooting the same subject with diverse settings and subsequently analyzing contact sheets, intuitively grasp the image dynamics. This cyclical practice of capturing, reviewing, and refining sharpens both their understanding of what constitutes a compelling image and the techniques to achieve it. A similar, albeit more pronounced, phenomenon occurs in AI image generation, where the inherently stochastic nature of the process might entail producing thousands of iterations for a single exhibit-worthy image. The analytical rigor this demands—differentiating the aesthetic merit of minor variations—is facilitated by automated contact sheets [reference the testing process using x/y/z grid automation].
  
  While the mediums differ, both photographers and AI artists hone their craft through a commitment to "learning by making", culminating in presenting selected pieces as finalized works. The parallel drawn between photography and AI art underscores the following aspects:
  - Rapid feedback loops.
  - The potential for extensive image production for comparative analysis.
  - An environment to discern and appreciate minute distinctions in images, thereby refining one's artistic sensibility.
  - A shorter learning curve to produce quality images compared to other art forms.
  
  Furthermore, mediums like photography and AI art challenge traditional notions of "artistic talent," often reserved for prowess in drawing, painting, or sculpting. This bias merits attention, especially considering the ongoing debates about AI art's legitimacy in both popular and scholarly circles [cite multiple references]. Such a biased view, albeit diminished since the rise of conceptual and media art, overlooks the significant exploration of art and creativity heralded in the late 19th century, epitomized by pioneers like DuChamp and later, figures like LeWitt.
  
  \subsection{Pedagogical Considerations within the Constructionist Framework}
  
  Building upon the discussion in the previous section, this segment aims to address key pedagogical considerations for teaching AI art tools.
  
  For artists to effectively utilize these tools to bring their imaginative visions to life, a foundational understanding of the basic concepts is essential. Drawing parallels with photography, it's noteworthy that a wide spectrum of technical proficiency can elevate an image's aesthetic quality. However, technical mastery isn't strictly necessary to produce aesthetically pleasing images. Esteemed photographers and artists such as Ansel Adams, Helmut Newton, Andy Warhol, and Robert Mapplethorpe often leveraged the simplicity of Polaroid cameras to profound artistic effect.
  
  Similarly, artists can produce original artworks without delving into the granular parameters and extensions described earlier, offering more refined control over images. This is evidenced by artists who predominantly employ tools like MidJourney through straightforward prompting. Notably, many of the foundational concepts of AI image generation software can be intuitively grasped through analogy or heuristics, rendering them accessible to those without a technical background.
  
  In line with the constructionist pedagogical approach, instruction should strike a balance: it should endow all students with a sufficient understanding of image generation, without inundating those less technically inclined with jargon or overly detailed descriptions. Given that these tools are open-source, those with a technical appetite can probe as deeply as desired. Open-source software's hallmark is its comprehensive documentation, offering insights into its construction.
  
  Furthermore, the expansive community of artists, spanning varied technical capabilities, ensures a wealth of learning resources are readily available. Embracing the tenets of constructionism necessitates a shift in the educator's role from a mere transmitter of knowledge to more of a guide or coach. This facilitates active learning through projects. Given the exploratory and experimental nature of AI art tools—which inherently promote the creation of novel artworks—constructionism arguably emerges as the ideal pedagogical approach for teaching their use.
  
  The subsequent section will delve into a series of introductory workshops on AI art creation using open-source software. It will feature three case studies spotlighting students at various stages of familiarity with AI tools. These students ventured further in their exploration, resulting in artworks that graced a group exhibition. Integral to this narrative will be an in-depth look at the teaching methods employed and their consequent outcomes.
  
  \subsection{Workshops, case studies and exhibition}
  In order to explore the feasibility of teaching open source software for AI image generation to university students with varying technical abilities it was decided to hold two workshops to teach the fundamentals of image creation with open-source AI tools.
  
  These workshops were held on July 19 and July 26 2023 and were open to undergraduates and graduates from any department at the university. Calls for participants were conducted through the posting of fliers around the university, word of mouth, and an announcement on the university’s digital bulletin system (Seika Portal). In total, 47 students from various departments enrolled in the workshops.
  
  \begin{figure}
      \centering
      \includegraphics[width=0.5\linewidth]{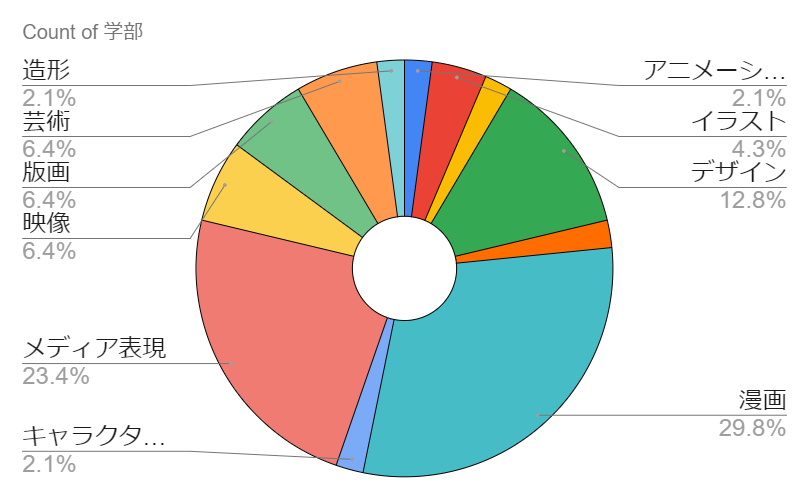}
      \caption{Enter Caption}
      \label{fig:enter-label}
  \end{figure}
  
  The workshops were designed such that students with no experience using these tools could participate in either the first workshop, the second workshop or both workshops. Additionally, online text and video tutorials were created and shared with participants as reference, and participants were invited to a private Discord server so that students could share their work, provide additional learning resources and receive technical support.
  
  At the end of each workshop a call was made for students to continue to explore AI tools in order to create works for an exhibition which was held from August 26th-September 3rd at the university. In total, three students from the workshops and one student who did not participate in the workshops created works for exhibition using the AI tools discussed here. The creation of exhibition ready images required a much greater degree of knowledge and technique, so the information presented at the workshops was expanded upon greatly over the following month by providing individual instruction and technical assistance to the exhibiting artists. A more detailed examination of this process is outlined below.
  
  \subsubsection{Workshops}
  When planning the workshops, the following assumptions were made:
  \begin{itemize}
      \item Students will primarily use Apple computers, eliminating the possibility of installing the software locally
      \item Students would have different desired outcomes that align roughly with their majors, i.e. Animation majors and Manga majors would want to create anime and manga style, fine art students would want to create works in original styles, etc.
      \item Students have no technical background and no experience of AI image tools.
  \end{itemize}

  In order to compensate for these, the following specifications were implemented in the design of the workshops:
  \begin{itemize}
  
      \item All of the software should run on the cloud and be operated through the browser. Additionally, shared cloud storage should be provided for students to save their generated images to.
      \item Techniques for working in different styles should be taught.
      \item Lessons should be planned such that no prior knowledge is required.
  
  \end{itemize}

  The most widely used user interface (UI) for Stable Diffusion is Automatic1111, which is a browser-based UI built with Gradio. As discussed above, there is a large community of active developers who regularly update the UI and integrate cutting edge research into the UI as extensions to the Stable Diffusion base. Although the UI is browser based, the standard implementation of it runs locally on the user’s computer and requires the actual processing of the images to be carried out on a graphics processing unit (GPU) with a certain amount of VRAM (6GB recommended minimum).  As it was assumed that the majority of students would be using Apple computers which do not meet these specifications, it was decided to use a cloud-based implementation of Automatic1111 that runs on a Google Colab notebook because Colab offers a limited amount of cloud processing for free to users.

  In order to allow students to explore creating with various styles of their choosing, a range of base models, VAE and LoRA were selected and uploaded to a shared Google Drive folder, and the students had full access to this shared drive to encourage them to upload their own models and LoRA for their testing and experimentation. Additionally, students created individual folders to in the shared drive for saving their output images.
  
  The contents of each lesson were adapted to text and video tutorials hosted on a shared Notion page as a resource for reference and review\footnote{https://www.notion.so/atticussims/Student-Page-AI-Art-Workshop-ab4a78199f924337824e0412185f3958}. 
  
  \paragraph{\textbf{Workshop 1: July 19, 2023}}
  
  In this inaugural workshop, the following topics were introduced:
  \begin{itemize}
  
    \item An introduction to AI image software, Stable Diffusion and the justifications for using open-source software
    \item An overview of the various resources used for open-source software, including Github, HuggingFace, CivitAI, Gradio \& Colab
    \item Initial setup and running of the Google Colab notebook and accessing the shared Google Drive
    \item An overview of Automatic1111 (Cagliostro WebUI version)
    \item Basic settings and key components of image generation, including models, prompts, seeds, samplers, steps, cfg, batches, and LoRA
    \item Testing and creating contact sheets with the X/Y/Z plot script
    \item Using the basic default upscaling tools. 
    \item Iterating images withImg2Img
  
  \end{itemize}
  
  \paragraph{\textbf{Workshop 2: July 26, 2023}}
  
  The goal of the second workshop was to provide students with an overview of the tools and extensions within Automatic1111 that allow for greater compositional and stylistic control of image generation. As a significant number of students who attended this workshop did not attend the first, it was also required to review the content of the first workshop. The following topics were introduced:
  \begin{itemize}
  
    \item Review of Workshop 1
    \item An introduction and overview of ControlNet with worked examples
    \item An introduction and overview of LoRA with worked examples.
  
  \end{itemize}
  
  \paragraph{Results of the workshops}
  Following the workshop, a survey was sent to all workshop participants requesting feedback about their perceptions of AI art, the tools introduced and their interest in attending future workshops or officially organized courses. Unfortunately, the number of respondents to the request were too few to present quantitative data in this paper.
  
  However, I would like to present some qualitative data based on my observations and from informal discussions with students which will influence the planning and conducting of future classes and workshops.
  
  \begin{enumerate}
  
    \item The setup and launching of the Google Colab notebook was complicated, and a number of students experienced trouble following the steps
  
    \item The time required to initiate the Automatic1111 UI was quite long due to limited cloud processing speed on free accounts and the requirement for downloading large files from the internet to the cloud drive.
  
    \item A certain percentage of participants received errors during the initiation process, causing frustration and a need to restart the process from the beginning.
  
    \item Once the technical difficulties were overcome, students generally had no problem using the software. Some students followed the instructions of the teacher precisely, while others immediately began experimenting and creating unique images.
  
    \item Many students expressed an interest in participating in future workshops.
  
  \end{enumerate}
  
  \paragraph{Analysis of Workshop Results}
  
  The feedback and observations from the workshop on AI image creation using open-source tools provide valuable insights into both the challenges faced by the students and the potential areas of improvement for the facilitators.

  \subparagraph{	Initial Technical Hurdles}
  
  \begin{itemize}
  
      \item A prominent challenge was the technicality involved in setting up and launching the Google Colab notebook. Students struggled with this initial step, highlighting a potential need for clearer instructions or a more user-friendly interface.
  
      \item The long initiation time of the Automatic1111 UI, compounded by the cloud processing speed constraints of free accounts and the need to download sizable files, further exacerbated the learning curve. This might have deterred some students or negatively impacted their enthusiasm in the early stages.
  
  \end{itemize}
  
  \subparagraph{	Errors and User Experience}
  
  Technical errors during the initiation process were encountered by a segment of the participants. The necessity to start over after encountering such issues can be a significant source of frustration. This indicates a potential need for refining the process, improving user guidance during this phase, or both.
  
  \subparagraph{	Positive Engagement Post-Technical Setup}
  
  \begin{itemize}
  
      \item Notably, after crossing the initial technical barriers, students seemed to navigate the software effectively. This suggests that the software, when running, is intuitive or that the instructions provided during the workshop were sufficient.
  
      \item It's worth noting the dichotomy in approach: some students adhered strictly to the workshop guidelines, while others opted for a more exploratory route. This highlights the diverse learning and creative styles of the participants, and future workshops might consider offering dual paths or flexible guidance to cater to both types of learners.
  
  \end{itemize}
  
  \subparagraph{	Interest in Continued Learning}
  
  Despite some of the initial technical challenges, there was a tangible interest among students to participate in subsequent workshops. This speaks to the inherent allure of AI image creation and suggests that if some of the initial barriers are addressed, engagement and satisfaction levels among participants could be even higher.

  \paragraph{Recommendations and Future Steps}
  Considering the feedback and observations, a few recommendations can be made:
  
  \begin{itemize}
  
    \item \textbf{Refinement of the Initial Setup}: Simplifying the setup process or providing additional guidance during the initial steps can reduce early-stage friction for students. This could be achieved by using a paid cloud computing service which would result in more, fewer steps for initiation, more efficient loading, and no need for downloading large files.
  
    \item \textbf{Technical Support}: Allocating a session or resources to troubleshoot common technical issues can help in smoother workshop experiences.
  
    \item \textbf{Continued Engagement}: The expressed interest in future workshops should be capitalized upon. Regular follow-ups, advanced sessions, or even creating a community space can foster continued learning and exploration in the field of AI art.
  
  \end{itemize}
  
  In summary, while the workshop faced some technical challenges, there's a clear appetite for learning in this domain. By addressing the initial hurdles and building on the positive engagement seen post-setup, future workshops can be even more effective and fulfilling for participants.

  \subsubsection{Learning by Making: A Collaborative Exploration}
  
  Following the described workshops, three participants, coming from distinct academic backgrounds and having varied prior experiences with AI, volunteered to produce artworks using AI tools. These works were showcased in a week-long group exhibition at Kyoto Seika University, commencing on 26 August 2023.
  
  \subsubsection{Student Profiles}
  
  \begin{enumerate}
      \item \textbf{Student 1}: 
      \begin{itemize}
          \item \textit{Affiliation}: A first-year master's student from the printmaking department.
          \item \textit{Background}: Lacked prior exposure to AI tools and had limited experience with digital art tools.
      \end{itemize}
      \item \textbf{Student 2}:
      \begin{itemize}
          \item \textit{Affiliation}: A fourth-year undergraduate from the printmaking department.
          \item \textit{Background}: Engaged in AI-related artworks for his graduation, familiar with MidJourney and ChatGPT, and had a basic understanding of Stable Diffusion and principles of diffusion models.
      \end{itemize}
      \item \textbf{Student 3}:
      \begin{itemize}
          \item \textit{Affiliation}: A third-year undergraduate majoring in architecture from the faculty of design.
          \item \textit{Background}: Proficient in AI art tools, recognized as a top architectural model creator on CivitAI. He showcased proficiency in creating and publishing various LoRA and custom models. Familiarity with Stable Diffusion was complemented by extensive work with MidJourney.
      \end{itemize}
  \end{enumerate}
  
  A fourth student, despite not attending the workshops but having individual AI art experience, also contributed to the exhibition. However, their works and experiences will not be examined in this paper.
  
  \subsection{Artistic Endeavors and Pedagogical Approaches}
  
  \begin{itemize}
      \item \textbf{Student 1}: Inspired by her emulsion wash silkscreen prints, the objective was to emulate this style in her AI creations. This necessitated the creation of a style-specific LoRA derived from her prints.
      \item \textbf{Student 2}: His exploration revolved around the confluence of human spirituality -- focusing on Buddhism and Shintoism -- with AI and the philosophical debate of machine spirituality. Drawing from conversations with ChatGPT on the topic, his vision was to produce images combining ancient Buddhist iconography with futuristic styles, encapsulating machine intelligence. To actualize this, he proposed training a LoRA grounded in imagery from ancient Japanese religious art.
      \item \textbf{Student 3}: With a vision to demonstrate the capabilities of AI in image creation, he aspired to amalgamate 200 diverse AI-generated images into one expansive artwork. Additionally, he displayed some of his architectural visualizations.
  \end{itemize}
  
  Given the technical intricacies involved in training a LoRA, technical support was extended to Students 1 and 2. To foster a deeper understanding of the process, these students were acquainted with the foundational concepts of LoRA training, with emphasis on curating an apt image dataset.
  
  As a testament to their expanded comprehension of AI art creation, a collaborative platform, a Miro Board, was instituted. This board facilitated contact sheet reviews, image annotations, feedback sharing, and iterative testing, enriching their learning experience.
  
  Enlargement of the finalized images for exhibit-quality prints presented a challenge. The native resolution of these images (768x1024 or 768x768 pixels) necessitated upscaling to at least 4K for optimal print quality. While Stable Diffusion offers basic image enlargement functionalities, achieving desired results required a nuanced approach. This involved leveraging a suite of extensions (ControlNet, Tiled Diffusion, and Ultimate SD Upscale), progressively doubling image size while retaining intricate details. Collaborative iterations ensured the final prints remained true to the artists' original visions, simultaneously deepening their grasp on digital imaging.

  \subsection{Analysis from a Constructionist Learning Perspective}
  
  \subsubsection{\textbf{Learning by Doing}:}
  
  The hands-on experience of the students clearly aligns with the constructionist principle of "learning by doing." \cite{Construc90:online} Rather than merely consuming knowledge, the students were actively involved in producing tangible artworks. This experiential learning allowed the students to deeply understand the AI tools in context, connecting their academic knowledge to real-world applications.
  
  \subsubsection{\textbf{Creation as Reflection of Understanding}:}
  
  As the students produced artworks, their creations became a reflection of their understanding. The nuances in their artistic choices, their interaction with the AI tools, and the final artworks all manifest their evolving comprehension of AI and its artistic possibilities. This aligns with the constructionist belief that when learners create something meaningful, it provides a tangible artifact of their understanding \cite{ackermann2001piaget}.
  
  \subsubsection{\textbf{Personalization of Learning}:}
  
  Each student brought their unique background and interests to the table, from emulsion wash silkscreen prints to explorations of spirituality and architectural visualizations. In constructionist learning, this personal context is pivotal. It aids in making the learning experience more engaging and ensures the knowledge gained is relevant to the individual learner's context\cite{kucirkova2019learning}.
  
  \subsubsection{\textbf{Iterative Process and Feedback}:}
  
  The iterative testing, use of the shared Miro Board for feedback, and the continuous refinement of their artworks mimic the iterative processes emphasized in constructionist pedagogy. Continuous feedback and iteration not only refine the end product but also deepen understanding and promote resilience in the face of challenges.
  
  \subsubsection{\textbf{Collaboration and Shared Learning}:}
  
  The collaborative nature of the learning process, evident in the shared board and the group exhibition, aligns with the constructionist perspective that learning is a social process. By collaborating, students could share insights, provide feedback, and learn from one another's experiences.
  
  \subsubsection{\textbf{Bridging the Knowledge Gap through Mentorship}:}
  
  The instructor's involvement in providing technical assistance, especially in the more complex areas like training the LoRA or enlarging images, exemplifies a scaffolding approach. This guided mentorship, where learners are supported in bridging their knowledge gaps, is a vital component of constructionist learning, allowing learners to undertake and succeed in tasks they might not manage alone \cite{wilson1996constructivist}.
  
  \subsubsection{\textbf{Emphasis on Understanding Over Memorization}:}
  
  By engaging with the foundational concepts of LoRA training and the nature of digital images, students were encouraged to prioritize deep understanding over rote memorization. This approach, where students understand the "why" behind processes, is central to constructionist pedagogy\cite{wilson1996constructivist}.
  
  \subsubsection{\textbf{Learning in a Real-world Context}:}
  
  The goal of creating artworks for an actual exhibition provides a real-world context to the learning process. Such authentic tasks make the learning meaningful and are a hallmark of constructionist pedagogy.
  
  In conclusion, the described experiences of students using AI tools to create artworks offers a compelling illustration of constructionist learning in action. The hands-on, iterative, and collaborative approach, combined with the emphasis on deep understanding and real-world application, embodies the core tenets of constructionist theory.
  
  \section{Discussion}
  
  It is widely believed that the current AI revolution is one of the most significant changes in human history. In a recent interview, AI pioneer Geoffrey Hinton stated that it is as important as the invention of the printing press or the invention of the wheel\cite{HintonInterview19:online}. I personally agree with this assessment and do not think it is an exaggeration in the least. 
  
  At the heart of this revolution are transformer based large language models (LLM) such as ChatGPT. These chat tools allow us to rapidly produce high quality content, build web pages and execute complex coding tasks.  LLMs in combination with image and audio synthesis AI frameworks have resulted in tools such as Stable Diffusion, Dall-E and Midjourney. These tools were only released in the second half of 2022, and already they are having a massive impact on every aspect of how we interact with and create media. This trend is further supported by my own research into the open-source generative AI ecosystem, where developers and artists are working together to expand the capabilities of foundation models, such as the Stable Diffusion model, by incorporating state of the art research in computer vision into low-code/no-code interfaces that enable non-technical artists use these tools.

  This trend will not slow down and the pace will only increase. GPT-4 was trained on the output of ChatGPT, and GPT-5 is rumored to be released in the near future with even greater accuracy and capabilities, including active access to the internet. The key idea is that this tool has created a feedback loop that quickens the pace of the trend, with people iteratively building better tools to build better tools and so on. The bottleneck of technical know-how has effectively been removed and with it a flood of innovative, high quality content. In essence, the barrier between imagination and product is diminishing to zero. We are at the beginning of an explosion of collective creative intelligence of the human species, and there is no going back. 
  
  Taking this into account, I believe that arts universities should fully embrace the use of AI. This includes teaching the use of AI in the classroom as well as using AI to improve the quality of our educational content and the efficiency of its delivery. If universities fail to adapt to the changes in industry, then we will not live up to one of our primary duties which is to prepare students for new jobs of the future. The AI revolution is happening incredibly fast, and if we do not take immediate action then we will be doing our current students a great disservice. By time this year’s freshmen graduate, creative industries will have fully adopted these technologies. In 5 years time, these industries be completely transformed, with new job titles, new tools, new product cycles and new marketing strategies. And, in all likelihood, far fewer new positions. If current students graduate without substantial generative AI skills, they will be competing for jobs with tens of thousands of people who are currently learning, utilizing and creating generative AI tools, and they will have difficulty attaining good positions in the job market of the near future. 
  This is why I strongly recommend that both university administration and individual teachers should take action now to begin incorporating these amazing, but incredibly disruptive new tools into classrooms and curricula. 
  
  \section{Conclusion}
  This study examined the significance of generative AI art tools in art universities. The discussion centered on AI's innovative potential and the impact of introducing generative technologies, particularly in image creation, on the landscape of contemporary art production. Through multiple workshops and exhibitions, the effectiveness and application methods of tools like Stable Diffusion in educational contexts were elucidated. While emphasizing the importance of collaboration between artists, developers, and researchers and highlighting AI's evolution in the artistic domain, the study also focused on challenges faced by educational institutions. Given the rapid integration of AI technology in the creative industry, it recognizes the importance of universities responding promptly to these changes. This movement symbolizes the intersection of tradition and revolution in art education and will likely serve as a strong driving force for incorporating advanced technological tools into educational curricula.

\end{document}